\documentclass[10pt,twocolumn,letterpaper]{article}

\usepackage{iccv}
\usepackage{times}
\usepackage{epsfig}
\usepackage{graphicx}
\usepackage{amsmath}
\usepackage{amssymb}
\usepackage[accsupp]{axessibility} 
\usepackage{bm}
\usepackage{algorithm}
\usepackage{algorithmic}
\usepackage{subfigure}
\usepackage{booktabs}
\usepackage{makecell}
\usepackage{multirow}
\usepackage{cuted}
\usepackage[colorlinks,linkcolor=red]{hyperref}

\iccvfinalcopy 

\def\iccvPaperID{7339} 

\ificcvfinal\fi

\begin{document}

\title{AutoDiffusion: Training-Free Optimization of Time Steps and Architectures for Automated Diffusion Model Acceleration}


\author{
    Lijiang Li$^{1}$,
    Huixia Li$^{2}$,
    Xiawu Zheng$^{1, 3, 4, 5}$,
    Jie Wu$^{2}$,
    Xuefeng Xiao$^{2}$,\\
    Rui Wang$^{2}$,
    Min Zheng$^{2}$,
    Xin Pan$^{2}$,
    Fei Chao$^{1}$\thanks{Corresponding author: fchao@xmu.edu.cn}, 
    Rongrong Ji$^{1, 3, 4, 5}$,\\
    $^1$Key Laboratory of Multimedia Trusted Perception and Efficient Computing,\\ Ministry of Education of China, Department of Artificial Intelligence, School of Informatics, \\
    Xiamen University. 
    $^2$ByteDance Inc. 
    $^3$Peng Cheng Laboratory. \\
    $^4$Institute of Artificial Intelligence, Xiamen University. 
    $^5$Fujian Engineering\\
    Research Center of Trusted Artificial Intelligence Analysis and Application, Xiamen University. \\
    {\tt\small lilijiang@stu.xmu.edu.cn, zhengxw01@pcl.ac.cn, wujie10558@gmail.com, \{feichao, rrji\}@xmu.edu.cn}\\
    {\tt\small \{lihuixia, xiaoxuefeng.ailab, ruiwang.rw, zhengmin.666, panxin.321\}@bytedance.com}
}

\maketitle
\ificcvfinal\thispagestyle{empty}\fi

\begin{abstract}
   Diffusion models are emerging expressive generative models, in which a large number of time steps (inference steps) are required for a single image generation. 
   To accelerate such tedious process, reducing steps uniformly is considered as an undisputed principle of diffusion models. 
   We consider that such a uniform assumption is not the optimal solution in practice; i.e., we can find different optimal time steps for different models. 
   Therefore, we propose to search the optimal time steps sequence and compressed model architecture in a unified framework to achieve effective image generation for diffusion models without any further training. 
   Specifically, we first design a unified search space that consists of all possible time steps and various architectures.
   Then, a two stage evolutionary algorithm is introduced to find the optimal solution in the designed search space. 
   To further accelerate the search process, we employ FID score between generated and real samples to estimate the performance of the sampled examples. 
   As a result, the proposed method is (\romannumeral1).\textbf{training-free}, obtaining the optimal time steps and model architecture without any training process; 
   (\romannumeral2). \textbf{orthogonal} to most advanced diffusion samplers and can be integrated to gain better sample quality. (\romannumeral3). \textbf{generalized}, where the searched time steps and architectures can be directly applied on different diffusion models with the same guidance scale. Experimental results show that our method achieves excellent performance by using only a few time steps, e.g. 17.86 FID score on ImageNet $64 \times 64$ with only four steps, compared to 138.66 with DDIM. The code is available at \href{https://github.com/lilijiangg/AutoDiffusion}{https://github.com/lilijiangg/AutoDiffusion}.
\end{abstract}

\section{Introduction}

\begin{figure*}
    \centering
    \includegraphics[width=0.95\textwidth]{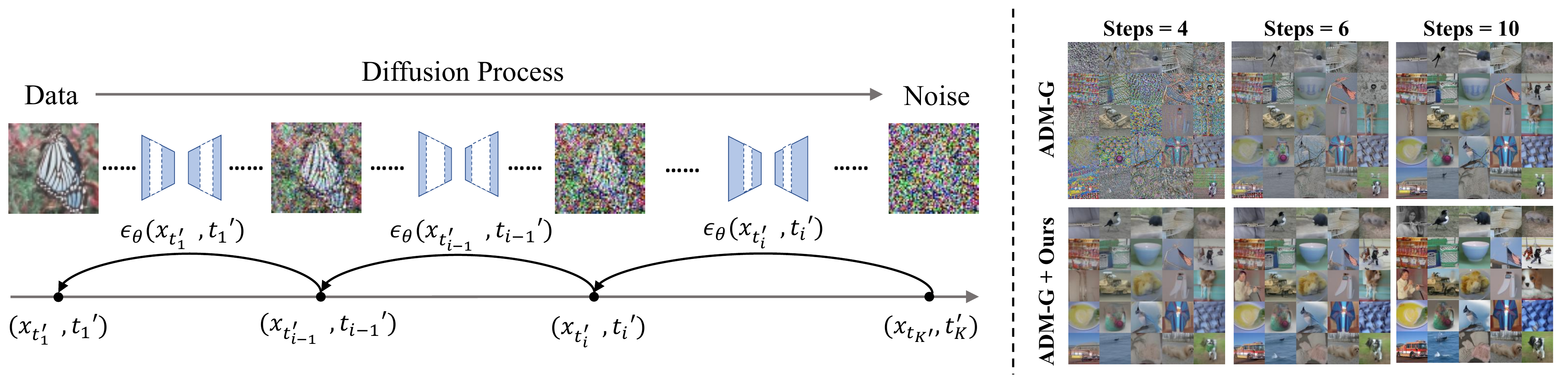}\label{Fig1}
    \caption{Left: We propose to search the optimal time steps sequence and corresponding compressed network architecture in a unified framework. Right: Samples by ADM-G \cite{ADM} pre-trained on ImageNet $64\times 64$ with and without our methods (AutoDiffusion), varying the number of time steps. 
    }
\vspace{-1.0em}
\end{figure*}

Diffusion models are a class of generative models that exhibit remarkable performance across a broad range of tasks, including but not limited to image generation \cite{ddpm, improved, ADM, analyticDPM, LDM, diffusion_img_generation, diffusion_img_generation1, sgm}, super-resolution \cite{sr3, SR_zero, sr_come_closer_faster}, inpainting \cite{inpaint, inpait1}, and text-to-image generation \cite{glide, Imagen, dall_e_2, txt2img1}. These models utilize the diffusion process to gradually introduce noise into the input data until it conforms to a Gaussian distribution. They then learn the reversal of this process to restore the data from sampled noise. Consequently, they achieve exact likelihood computation and excellent sample quality. However, one major drawback of diffusion models is their slow generation process. For instance, on a V100 GPU, generating a $256 \times 256$ image with StyleGAN \cite{stylegan2} only takes 0.015s, whereas the ADM model \cite{ADM} requires multiple time steps for denoising during generation, leading to a significantly longer generation time of 14.75s.

Extensive studies have focused on reducing the number of time steps to improve the generation process of diffusion models. Some of these studies represent the generation process as either stochastic differential equations (SDEs) or ordinary differential equations (ODEs), and then utilize numerical methods to solve these equations \cite{ddim, plms, sr_come_closer_faster, dpmsolver}. The samplers obtained by these numerical methods can typically be applied to pre-trained diffusion models in a plug-and-play manner without re-training. The other studies proposed to utilize knowledge distillation to reduce the number of time steps \cite{progressive_kd, kd1}. These methods decrease the time steps required for the generation process and then allow the noise prediction network to learn from the network of the original generation process. Although these methods are effective in improving the sampling speed of diffusion models, we observe that they have paid little attention to the selection of time step sequences. When reducing the number of time steps, most of these methods sample the new time steps uniformly or according to a specific procedure \cite{ddim}. We argue that there exists an optimal time steps sequence with any given length for the given diffusion model. And the optimal time steps sequence varies depending on the specific task and the super-parameters of diffusion models. We believe that the generation quality of diffusion models can be improved by replacing the original time steps with the optimal time steps.

Therefore, we introduce AutoDiffusion, a novel framework that simultaneously searches optimal time step sequences and the architectures for pre-trained diffusion models without additional training. 
Fig.~\ref{Fig1} (Left) shows the schematic of AutoDiffusion. Our approach is inspired by Neural Architecture Search (NAS) techniques that are widely used for compressing large-scale neural networks \cite{nas1, nas2, nas5, nas6, nas7}. In our method, we begin with a pre-trained diffusion model and a desired number of time steps. 
Next, we construct a unified search space comprising all possible time step sequences and diverse noise prediction network architectures. 
To explore the search space effectively, we use the distance between generated and real samples as the evaluation metric to estimate performance for candidate time steps and architectures. Our method provides three main advantages. First, we demonstrate through experiments that the optimal time steps sequence obtained through our approach leads to significantly better image quality than uniform time steps, especially in a few-step regime, as illustrated in Fig.~\ref{Fig1} (Right). 
Second, we show that the searched result of the diffusion model can be applied to another model using the same guidance scale without repeating the search process. 
Furthermore, our approach can be combined with existing advanced samplers to further improve sample quality.

Our main contributions are summarized as follows:

\begin{itemize}
    \item Our study reveals that uniform sampling or using a fixed function to sample time steps is suboptimal for diffusion models. Instead, we propose that there exist an optimal time steps sequence and corresponding noise prediction network architecture for each diffusion model. To facilitate this, we propose a search space that encompasses both time steps and network architectures. Employing the optimal candidate of this search space can effectively improve sampling speed for diffusion models and complement the most advanced samplers to enhance sample quality. 
    \item We propose a unified training-free framework, AutoDiffusion, to search both time steps and architectures in the search space for any given diffusion model. We utilize a two-stage evolutionary algorithm as a search strategy and the FID score as the performance estimation for candidates in the search space, enabling an efficient and effective search process. 
    \item Extensive experiments show that our method is training-free, orthogonal to most advanced diffusion samplers, and generalized, where the searched time steps and architectures can be directly applied to different diffusion models with the same guidance scale. Our method achieves excellent performance by using only a few time steps, \emph{e.g.}, 17.86 FID score on ImageNet $64 \times 64$ with only four steps, compared to 138.66 with DDIM. Furthermore, by implementing our method, the samplers exhibit a noteworthy enhancement in generation speed, achieving a $2 \times$ speedup compared to the samplers lacking our method. 
\end{itemize}

\begin{figure*}[h!]
    \centering
    \includegraphics[width=1.0\textwidth]{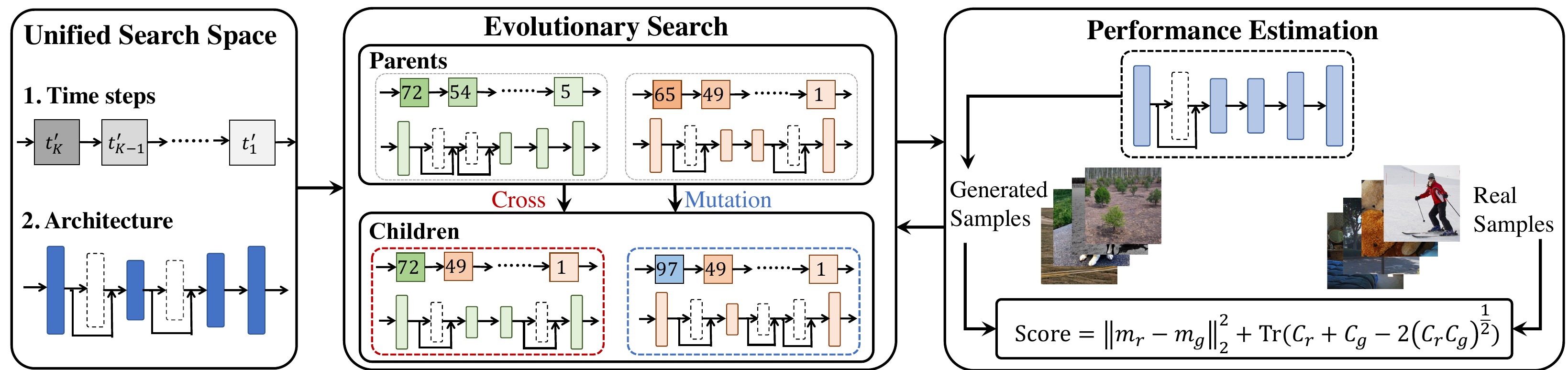}
    \caption{Overview on AutoDiffusion. Given a pre-trained diffusion model, we first design a unified search space that consists of both time steps and architectures. After that, we utilize the FID score as the performance estimation strategy. Finally, we apply the evolutionary algorithm to search for the optimal time steps sequence and architecture in the unified search space.}
    \label{Fig_framework}
\vspace{-1.0em}
\end{figure*}
\section{Related Work}
\subsection{Diffusion Models}
Given a variable $x_0 \in \mathbb{R}^D$ that sampled from an unknown distribution $p_{data}(x_0)$, diffusion models define a diffusion process $\{x_t\}_{t\in[0:T]}$ to convert the data $x_0$ into sample $x_T$ by $T$ diffusion steps. The distribution of the sample $x_T$ denoted as $p(x_T)$ is usually simple and tractable, such as standard normal distribution. In the diffusion process, the distribution of variable $x_t$ at time step $t$ satisfies:
\begin{equation}
\label{q_xt_x0}
    q(x_t|x_0) = \mathcal{N}(x_t|\alpha_t x_0, \beta^2_tI)
\end{equation}
where $\{\alpha_1, \alpha_2, \cdots, \alpha_T\}$ and $\{\beta_1, \beta_2, \cdots, \beta_T\}$ are super-parameters of diffusion models that control the speed of converting $x_0$ into $x_T$. 

After that, diffusion models define a reverse process $p_\theta(x_{t-1}|x_t)$ parameterized by neural network $\theta$ and optimize it by maximizing the log evidence lower bound (ELBO) \cite{improved}:
\begin{equation}
\begin{split}
    \label{l_elbo}
    L_{elbo} &= \mathbb{E}[\log p_\theta(x_0|x_1) \\ &- \Sigma_{t=1}^T D_{KL}(q(x_{t-1}|x_t, x_0)||p_\theta(x_{t-1}|x_t)) \\&- D_{KL}(q(x_T|x_0)||p(x_T))]
\end{split}
\end{equation}
where $D_{KL}$ denote the KL-divergence.

In practice, diffusion models use a noise prediction network $\epsilon_\theta(x_t, t)$ to estimate the noise component of the noisy sample $x_t$ at time step $t$. Therefore, the loss function in Eq.~\ref{l_elbo} can be simplified as follow \cite{ddpm}:
\begin{equation}
    \label{l_simple}
    L_{simple} = \left\| \epsilon_\theta(x_t, t) - \epsilon\right\|^2
\end{equation}
where $\epsilon$ represent the noise component of $x_t$ and we have $x_t = \alpha_t x_0 + \beta \epsilon$ according to Eq.~\ref{q_xt_x0}. In most diffusion models, the noise $\epsilon$ is sampled from standard normal distribution $\mathcal{N}(0, I)$ when generating noisy sample $x_t$.

When the noise prediction network $\epsilon_\theta(x_t, t)$ is trained, diffusion models define a generation process to obtain samples. 
This process begins with noisy data sampled from $p(x_T)$, yielding progressively cleaner samples $x_{T-1}, x_{T-2}, \cdots, x_0$ via the learned distribution $p_\theta(x_{t-1}|x_t)$. This process needs $T$ forward of the noise prediction network $\epsilon_\theta$ to obtain final sample $x_0$. To hasten this, many studies tried to reduce the number of time steps to $K<T$. They proposed many advanced samplers to compensate for the loss of sample quality caused by reducing time steps. But most of them overlooked optimal time step selection and usually sampled new time steps based on simple functions. For example, DDIM \cite{ddim} select time steps in the linear or quadratic procedure. The linear procedure generate new time steps sequence with length $K$ such that $[0, \frac{T}{K}, \cdots, \frac{KT}{K}]$. Our key contribution is searching the $K$-length optimal time steps sequence for diffusion models.  

\subsection{Neural Architecture Search}
The aim of NAS algorithms is to automatically search for an appropriate neural network architecture within an extensive search space. NAS is composed of three essential components: the search space, the search strategy, and the performance estimation strategy \cite{nas_overview}. The search space specifies the set of architectures to be explored and determines the representation of candidate neural networks. The search strategy outlines the approach employed to explore the search space. Typically, the strategy involves selecting a new candidate from the search space based on the performance estimation of the currently selected candidate. The performance estimation strategy defines the approach for evaluating the performance of a candidate neural network in the search space. An effective performance estimation strategy ensures accurate and swift evaluations, underpinning both the efficacy and speed of the NAS \cite{pe}.

NAS algorithms have been applied to design suitable network architecture in various fields. Therefore, in this work, we aim to optimize the time steps and architecture of diffusion models using this technique.

\subsection{Fast Sampling For Diffusion Models}
Numerous studies aim to improve the generation speed of diffusion models. Some approaches model the generation process with SDEs or ODEs, leading to training-free samplers \cite{ddim, plms, dpmsolver}. However, when the number of steps drops below 10, these methods often degrade image quality \cite{TRACT}. Other methods accelerate diffusion models via knowledge distillation \cite{progressive_kd, kd1, TRACT} or by learning a fast sampler \cite{DDSS}. For example, progressive distillation (PD) uses knowledge distillation to halve the number of time steps \cite{progressive_kd}. This distillation is iteratively conducted until the number of steps is less than 10, often demanding substantial computational resources. DDSS treats sampler design as a differentiable optimization problem, utilizing the reparametrization trick and gradient rematerialization to learn a fast sampler \cite{DDSS}. Although DDSS offers notable speedups, it lacks flexibility, as samplers tailored for one model may not fit another, requiring distinct learning stages. Compared with these methods, AutoDiffusion is much more efficient and flexible, as substantiated by our experiments. Its searched result can be transferred to another diffusion model using the same guidance scale without re-searching. Furthermore, AutoDiffusion utilizes a unified search space for time steps and model layers, while existing methods only focus on step reduction.

\section{Method}
In this section, we introduce our AutoDiffusion, which aims to search for the optimal time steps sequence and architecture for given diffusion models. The overview of our method is shown in Fig.~\ref{Fig_framework}. In the following contents, we first discuss the motivation of our method in Sec.~\ref{motivation}. Then, we introduce the search space in Sec.~\ref{search_space}. After that, we elaborate the performance evaluation in Sec. \ref{pe}. Finally, we introduce the evolutionary search in Sec. \ref{ea}. 


\subsection{Motivation}
\label{motivation}


Many well-recognized theories pointed out that the generation process of diffusion models is divided into several stages, in which the behavior of diffusion models is different at each stage \cite{three_stage, multi_stage}. For example, Ref \cite{three_stage} illustrated that the behavior of diffusion models at each time step can be classified into creating coarse features, generating perceptually rich contents, and removing remaining noise. Intuitively, the difficulty of these tasks is different. In other words, the denoise difficulty of diffusion models varies with the time steps. 
Inspired by these studies, we hypothesize that the importance of each time step in the generation process is different. In this case, we argue that there exists an optimal time steps sequence for diffusion models among all possible time steps sequences. 

\begin{figure}
    \centering
    \includegraphics[width=0.48\textwidth]{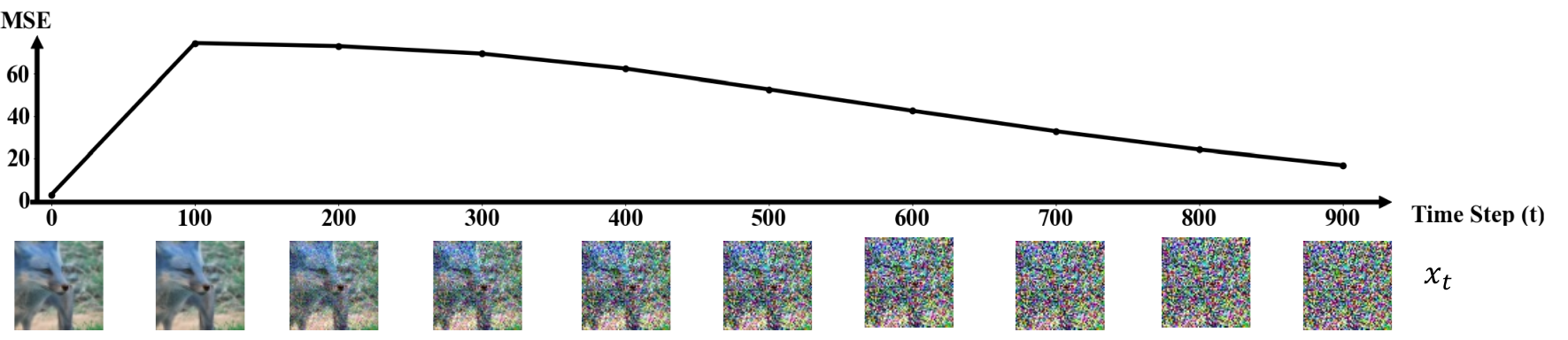}
    \caption{Sample $x_t$ and MSE $\left\|x_t - x_{t+100}\right\|^2$ over time steps $t$. }
    \label{Fig_timesteps}
\vspace{-1.0em}
\end{figure}

To investigate our hypothesis, we conduct an experiment in which we obtain samples, denoted as $x_t$, and calculate the Mean Squared Error (MSE) $\left\|x_t - x_{t+100}\right\|^2$ for each time step $t$. The results are presented in Fig. \ref{Fig_timesteps}, which shows that the samples obtained for $t \in [600, 1000]$ are dominated by noise and thus illegible. Conversely, when $t \in [300, 600]$, the diffusion model generated the main contents of the image, and the objects in the generated image become recognizable. It is observed that the diffusion model primarily removes noise at $t \in [0, 300]$, resulting in similar samples for $t \in [0, 300]$. Furthermore, Fig.~\ref{Fig_timesteps} indicates that the MSE is low at $t \in [0, 100]$ and $t \in [700, 900]$, while it becomes high at $t \in [200, 600]$. Based on the findings in Fig.~\ref{Fig_timesteps}, it is apparent that different time steps play varying roles in the generation process of diffusion models. Specifically, when $t$ is small or large, the content of the generated samples changes slowly. In contrast, when $t$ is in the middle, the content changes rapidly. Therefore, we contend that uniform time steps are suboptimal and that an optimal time step sequence exists for the generation process of diffusion models. Further, since the denoise difficulty varies depending on time steps, we believe that the model size of the noise prediction network is not necessarily the same at each time step. Thus, we search the time steps and architectures in a unified framework. 
\subsection{Search Space}
\label{search_space}
In this section, we discuss how the search space is designed in AutoDiffusion. 
Given a diffusion model with timesteps $[t_1, t_2, \cdots, t_T]$ ($t_i < t_{i+1}$), it needs $T$ calls of the noise prediction network $\epsilon_\theta$ to yield a batch of images. To accelerate the generation process, two approaches are usually employed: reducing the number of time steps or the number of layers in the $\epsilon_\theta$. To this end, we propose a search space comprising two orthogonal components: 1) the temporal search space that takes time steps as the searched object; 2) the spatial search space that takes the architectures of the noise prediction network $\epsilon_\theta$ as the searched object. In our search space, the candidate $cand$ is defined as follows:
\begin{equation}
\begin{split}
    cand &= \{ \bm{\mathcal{T}} = [t_1', t_2', \cdots, t_K']; \\
        & \bm{\mathcal{L}} = [\bm{L_1}, \bm{L_2}, \cdots, \bm{L_K}] 
        \}, \\
                & 0 < t_{i+1}' - t_i' < t_T - t_1, \\
                & t_i' \in [t_1, t_2, \cdots, t_T] ~ (i = 1, 2, \cdots, K)
\label{Eq:search_space}
\end{split}
\end{equation}
where $\bm{\mathcal{T}}$ denotes the sampled time steps sequence, and $[t_1', t_2', \cdots, t_K']$ is a sub-sequence of the original time steps sequence $[t_1, t_2, \cdots, t_T]$. $\bm{\mathcal{L}}$ denotes the sampled architectures, where $\bm{L_i} = [l_i^1, l_i^2, \cdots, l_i^{n_i}]$ is the architecture of the noise prediction model at time step $t_i'$. $n_i$ is the number of architecture layers at time step $t_i'$ , which must be no more than the layers number of $\epsilon_\theta$. 
Each $l_i^j \in \bm{L_i}$ represents one layer of the noise prediction network $\epsilon_\theta$ at time step $t_i'$, thus $\bm{L_i}$ can be viewed as a sub-network of $\epsilon_\theta$.
In practice, we constrain the sum of model layers at each time step to be no more than $N_{\text{max}}$, \emph{i.e.} $\sum_{i=1}^K n_i \leq N_{\text{max}}$, where $N_{\text{max}}$ is determined according to the expected generation speed of diffusion models.

In the temporal aspect, we search for the optimal time steps sequence among all possible time steps. In the spatial aspect, we search for the model layers of the noise prediction network at each time step. Therefore, we can search for the best time steps sequence and the compressed noise prediction model in a unified framework. Notably, the sub-network $\bm{L_i}$ may not be the same across all time steps during the search, as the difficulty of denoising varies at different time steps. We believe that the number of layers $n_i$ at each time step $t_i'$ reflects the level of denoising difficulty at $t_i'$.

Since the noise prediction networks $\epsilon_\theta$ are usually U-Net, we don't add up-sample or down-sample layers into the search space. In practice, if a model layer is not selected in a candidate, the model layer will be replaced by a skip connection. Besides, the searched sub-networks of $\epsilon_\theta$ are not retrained or fine-tuned in the search process.

\subsection{Performance Estimation}
\label{pe}
After the search space is determined, we need to select evaluation metrics to provide fast and proper performance estimation for the search process. There are two classes of evaluation metrics that may meet the requirements, one is the distance between learned distribution $p_\theta(x_{t_{i-1}}|x_{t_i})$ and posteriors $q(x_{t_{i-1}}|x_{t_i}, x_0)$, the other is the distance between the statistics of generated samples and real samples. 

The distance between distribution $p_\theta(x_{t_{i-1}}|x_{t_i})$ and posteriors $q(x_{t_{i-1}}|x_{t_i}, x_0)$ is usually estimated using KL-divergence. Therefore, the performance estimation of a sorted candidate time steps $[t_1', t_2', \cdots, t_K']$ can be obtained by using KL-divergence \cite{improved} as follows:
\begin{equation}
\begin{split}
    \label{KL}
    L &= L_{t_1'} + L_{t_2'} + \cdots + L_{t_K'} \\
    L_{t_i'} &= 
    \begin{cases}
        D_{KL}(q(x_{t_i'}|x_0)||p(x_{t_i'})),\quad & t_i' = t_T \\
        -\log p_\theta(x_{t_i'}|x_{t_{i+1}'}),\quad & t_i' = 0 \\
        D_{KL}(q(x_{t_i'}|x_{t_{i+1}'}, x_0)||p_\theta(x_{t_i'}|x_{t_{i+1}'}),\quad & others
    \end{cases}
\end{split}
\end{equation}

Given a trained diffusion model, the image $x_0$ sampled from the training dataset, and the candidate time steps $[t_1', t_2', \cdots, t_K']$, we use Eq.~\ref{KL} to calculate the KL divergence, which allows a fast performance estimation. However, prior work has pointed out that optimizing the KL-divergence can not improve sample quality \cite{learned_sampler, learned_sampler_ref}. 
To verify this conclusion, we use the time steps sequence $[t_1, t_2, \cdots, t_T]$ of a diffusion model trained on ImageNet $64 \times 64$ as the search space. Then, we sample subsequences $[t_1', t_2', \cdots, t_K']$ from this search space randomly and calculate the FID score, sFID score, IS score, precision, recall, and the KL-divergence of these subsequences. After that, we analyze the relevancy between FID, sFID, IS, precision, recall, and KL-divergence of these subsequences by calculating the Kendall-tau \cite{kendall} between them. Tab.~\ref{kdt} shows that the Kendall-tau values between all these metrics and KL-divergence are low, which means that the KL-divergence can not represent the sampled quality. 

\begin{table}
\begin{center}
\begin{tabular}{c|c|c|c|c}
\toprule
FID & sFID & IS & Precision & Recall \\
\hline
0.126 & 0.200 & -0.126 & -0.190 & -0.165\\
\bottomrule
\end{tabular}
\end{center}
\vspace{-0.8em}
\caption{Kendall-tau \cite{kendall} between matrices and KL-divergence.}
\vspace{-1.5em}
\label{kdt}
\end{table}


The distance between the statistics of generated samples and real samples can be estimated using the KID score or FID score. 
Daniel~\emph{et al.} proposed to optimize the sampler of diffusion models by minimizing KID loss \cite{DDSS}. Inspired by this work, we use FID score as the performance estimation metric. The FID score is formulated as follows \cite{fid}:
\begin{equation}
    \text{Score} = \left\| m_r - m_g\right\|^2_2 + \text{Tr}\left(C_r+C_g-2(C_rC_g)^{\frac{1}{2}}\right)
\label{eq_fid}
\end{equation}
where $m_r$ and $m_g$ are the mean of the feature of real samples and generated samples; while $C_r$ and $C_g$ are covariances of the feature of real samples and generated samples. Usually, the feature of generated samples and real samples can be obtained by pretrained VGG \cite{vgg} models. 

However, we must generate at least 10k samples when calculating precise FID scores, which will slow down the search speed. To address this, we reduce the number of samples for calculating FID scores. We apply Kendall-tau \cite{kendall} to determine the reduced number of samples. Specifically, we still use the full time steps sequence $[t_1, t_2, \cdots, t_T]$ as search space and sample $N_{seq}$ subsequences $[t_1', t_2', \cdots, t_K']$ randomly from it. Then, we generate 50k samples using each of these subsequences and obtain corresponding FID scores $\{F_1, F_2, \cdots, F_{N_{seq}}\}$. After that, we obtain a subset of $N_{sam}$ samples from 50k samples and calculate their FID score $\{F_1', F_2', \cdots, F_{N_{seq}}'\}$. We calculate the Kendall-tau between $\{F_1, F_2, \cdots, F_{N_{seq}}\}$ and $\{F_1', F_2', \cdots, F_{N_{seq}}'\}$. The optimal number of samples is the minimum $N_{sam}$ that makes Kendall-tau greater than 0.5. 

\subsection{Evolutionary Search}
\label{ea}
We utilize the evolution algorithm to search for the best candidate from the search space since evolutionary search is widely adopted in previous NAS works\cite{nas1, nas_ea_1, nas_ea_2, nas_ea_3}. 
In the evolutionary search process, given a trained diffusion model, we sample candidates from the search space randomly using Eq.~\ref{Eq:search_space} to form an initial population. For each candidate, we generate samples by utilizing the candidate's time steps and corresponding architecture. After that, we calculate the FID score based on the generated samples. At each iteration, we select the Top $k$ candidates with the lowest FID score as parents and apply cross and mutation to generate a new population. To perform cross, we randomly exchange the time steps and model layers between two parent candidates. To perform mutation, we choose a parent candidate and modify its time steps and model layers with probability $p$. 

When searching for time steps and architectures, we utilize a two-stage evolutionary search. Specifically, we use the full noise prediction network and search time steps only in the first several iterations of the evolutionary search. Then, we search the time steps and model architectures together in the remaining search process. 
    
\section{Experimentation}

\subsection{Experiment Setting}
In order to demonstrate that our method is compatible with any pre-trained diffusion models, we apply our method to prior proposed diffusion models. Specifically, we experiment with the ADM and ADM-G models proposed by Prafulla~\emph{et al.}\cite{ADM} that trained on ImageNet $64\times64$ \cite{imagenet} and LSUN dataset \cite{lsun}. In addition, we applied our method on Stable Diffusion \cite{LDM} to verify the effectiveness of our method on the text-to-image generation task. Besides, we also combine our method with DDIM \cite{ddim}, PLMS \cite{plms}, and DPM-solver \cite{dpmsolver} and apply them to the Stable Diffusion 
to demonstrate that our proposed method can be combined with most of the existing advanced samplers and improve their performance. In all experiments, we use the pre-trained checkpoint of these prior works since our method does not need to retrain or fine-tune the diffusion models. 

Our method optimizes the generation process of diffusion models from the perspective of both time steps and architecture. Sec.~\ref{sec.main_results} illustrates that we can accelerate the generation process by only searching for the optimal time steps. And on this basis, Sec.~\ref{sec.search_timesteps_and_arch} demonstrates that we can improve the sample quality and generation speed further by searching time steps and architecture together. 
In all experiments, the hyperparameters of evolution algorithm search are set as follows: we set the population size $P = 50$; top number $k = 10$, mutation probability $p = 0.25$, max iterations $MaxIter = 10$ when searching for time steps only, and $MaxIter = 15$ when searching for time steps and architectures. For the experiments without our methods, the diffusion models generate samples with uniform time steps and the full noise prediction network. Besides, all experiments with ADM or ADM-G use DDIM \cite{ddim} sampler. We evaluate the quality of generated images with FID and IS scores as most previous work.  

\subsection{Quantitative and Qualitative Results}
\label{sec.main_results}
We apply our method with the pre-trained ADM-G and ADM on various datasets, and the results are shown in Tabs.~\ref{main_results_ImageNet64} to \ref{main_results_LSUN}. Note that we only search time steps without searching model layers of the noise prediction network in these experiments. Our method can improve the sample quality significantly of diffusion models in the few-step regime. In particular, our method exhibits impressive performance when the number of time steps is extremely low. For example, the FID score of ADM-G on ImageNet $64\times 64$ is 138.66, and our method can reduce it to 17.86, which shows that our method can generate good samples in the extremely low-step regime. 

\begin{table}
    \begin{center}
    \begin{tabular}{cccc}
    \toprule[1pt]
    Ours & Steps & FID $\downarrow$ & IS $\uparrow$ \\
    \hline
    $\times$  & 4 & 138.66 & 7.09 \\
    $\checkmark$ & 4 & 17.86 (-120.8) & 34.88 (+27.79) \\
    \hline
    $\times$  & 6 & 23.71 & 31.53 \\
    $\checkmark$   & 6 & 11.17 (-12.54)  & 43.47 (+11.94) \\
    \hline
    $\times$  & 10 & 8.86 & 46.50 \\
    $\checkmark$   & 10 & 6.24 (-2.62) & 57.85 (+11.35) \\
    \hline
    $\times$  & 15 & 5.38 & 54.82 \\
    $\checkmark$   & 15 & 4.92 (-0.46) & 64.03 (+9.21)\\
    \hline
    $\times$  & 20 & 4.35 & 58.41 \\
    $\checkmark$   & 20 & 3.93 (-0.42) & 68.05 (+9.64) \\
    \toprule[1pt]
    \end{tabular}
    \end{center}
    \caption{FID ($\downarrow$) and IS ($\uparrow$) scores for ADM-G\cite{ADM} with and without our method on ImageNet $64\times 64$, varying the number of time steps. The (+number) denotes the improve compare to the resulte without our method.}
\label{main_results_ImageNet64}
\end{table}

\begin{table}
    \begin{center}
    \begin{tabular}{cccc}
    \toprule[1pt]
    Ours & Steps & LSUN Bedroom & LSUN Cat\\
    \hline
    $\times$  & 5 & 33.42  & 48.41 \\
    $\checkmark$ & 5 & 23.19 (-10.23) &  34.61 (-13.8) \\
    \hline
    $\times$  & 10 & 10.01 & 20.05 \\
    $\checkmark$ & 10  & 8.66 (-1.35) & 17.29 (-2.76) \\
    \hline
    $\times$  & 15 & 6.36 & 14.86 \\
    $\checkmark$   & 15 & 5.83 (-0.53) & 13.17 (-1.69) \\
    \toprule[1pt]
    \end{tabular}
    \end{center}
    \caption{FID score ($\downarrow$) for ADM\cite{ADM} with and without our method on LSUN dataset, varying the number of time steps.}
\label{main_results_LSUN}
\vspace{-0.5cm}
\end{table}
\begin{figure}
\begin{center}
\includegraphics[width=0.45\textwidth]{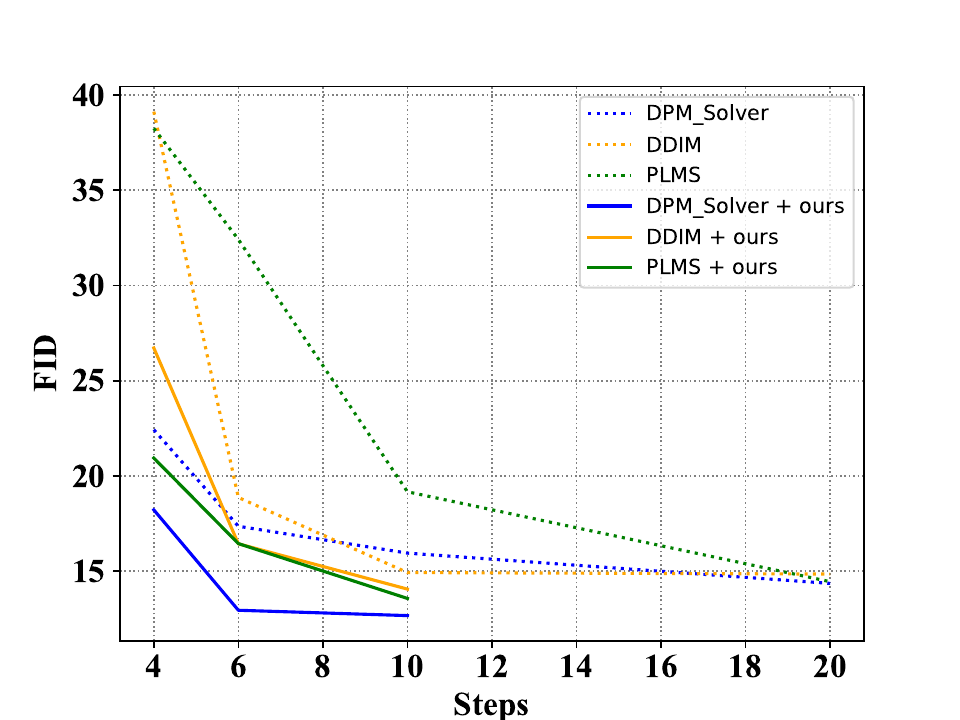}
\end{center}
\vspace{-0.3cm}
   \caption{FID score for Stable Diffusion \cite{LDM} using different samplers with and without our methods. Our method can improve the FID score of DDIM, PLMS, and DPM-solver. }
\label{fig.sd_timesteps}
\vspace{-0.5cm}
\end{figure}

\begin{figure*}
\begin{center}
\includegraphics[width=0.90\textwidth]{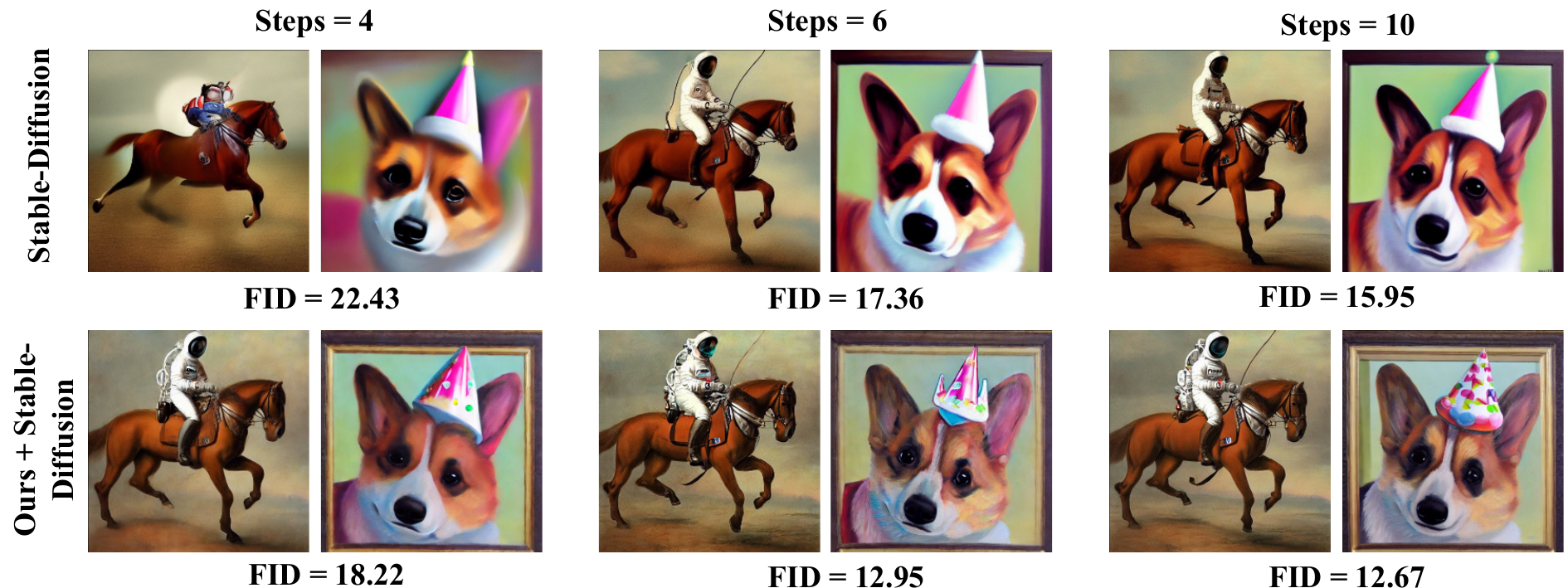}
\end{center}
\vspace{-1.0em}
   \caption{Samples by Stable diffusion \cite{LDM} with and without our methods using the same random seed, varying the number of time steps. Input prompts are ``An astronaut riding a horse'' and ``An oil painting of a corgi wearing a party hat''.}
\label{fig.sd_samples}
\end{figure*}
\begin{figure*}
\begin{center}
\includegraphics[width=0.95\textwidth]{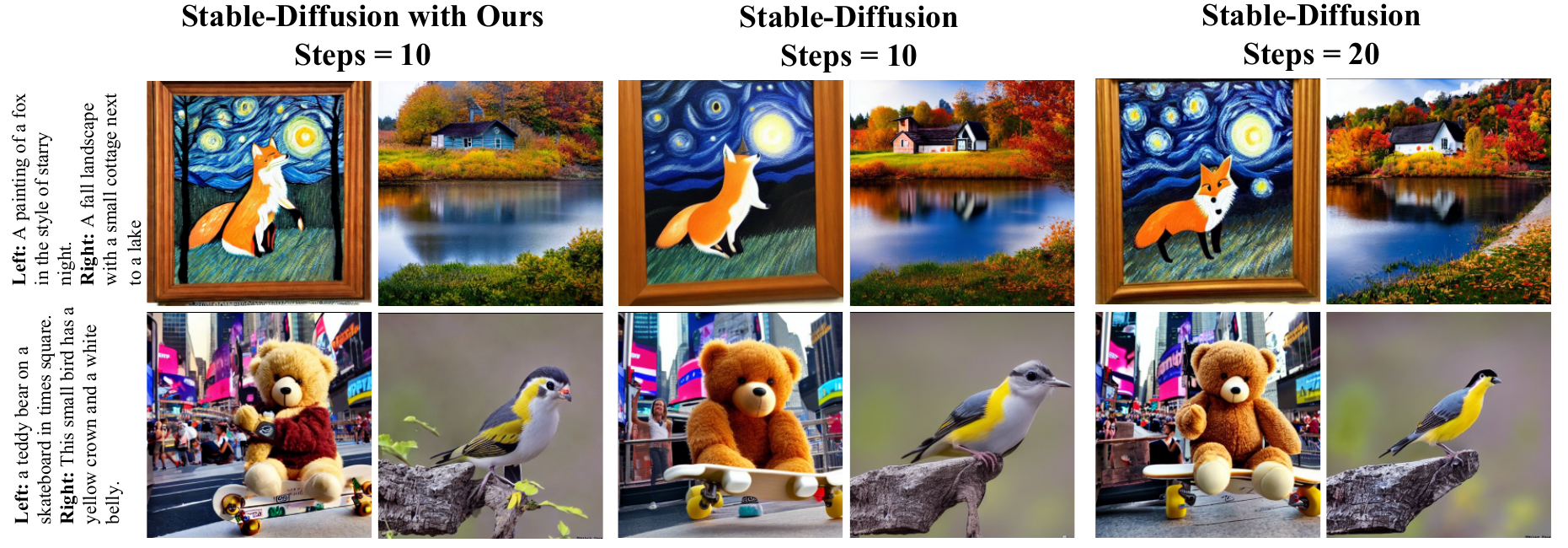}
\end{center}
\vspace{-1.0em}
   \caption{The proposed method is also compatible with the widely-used sampler  DPM-Solver. The samples generated by our method with 10 steps are comparable to those generated by 20 steps, and better than those generated by 10 steps using DPM-Solver.}
\label{fig.sd_samples_part2}
\vspace{-1.2em}
\end{figure*}

We combine our method with DPM-Solver \cite{dpmsolver}, DDIM \cite{ddim}, and PLMS \cite{plms} to demonstrate that our method can be integrated with advanced samplers. 
Fig.~\ref{fig.sd_timesteps} shows that our method can improve the sample quality based on these samplers, especially in the low-step case where steps = 4. These results illustrate that our method can be combined with most advanced samplers to further improve their performance. In addition, Fig.~\ref{fig.sd_timesteps} illustrates that the samplers with our method can achieve admirable performance within 10 steps, which is $2\times$ faster than the samplers without our method. 

Fig.~\ref{fig.sd_samples} shows the generated samples for Stable diffusion using DPM-Solver with and without our method in a few-step regime. We find that the samples generated with our method have more clear details than other samples. Fig.~\ref{fig.sd_samples_part2} demonstrates that the images generated by our method with DPM-Solver at step = 10 are comparable to those generated solely by DPM-Solver at step = 20, and superior to those generated solely by DPM-Solver at step = 10. 

\begin{figure}
    \centering
    \subfigure[]{\includegraphics[width=0.235\textwidth]{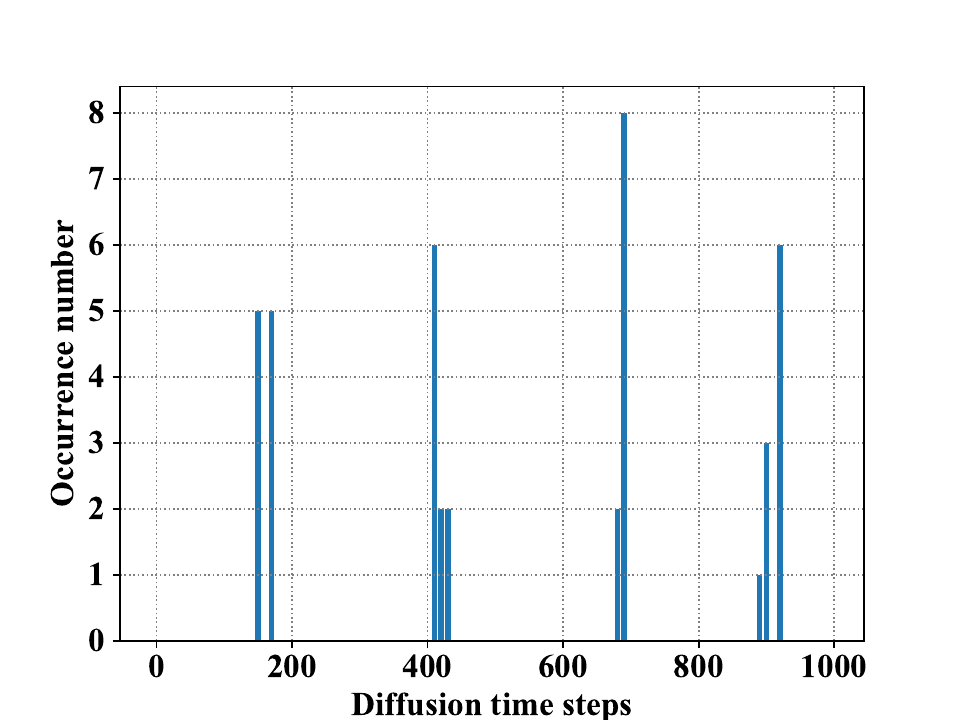}\label{Fig.migrate_adm_4}}
    \subfigure[]{\includegraphics[width=0.235\textwidth]{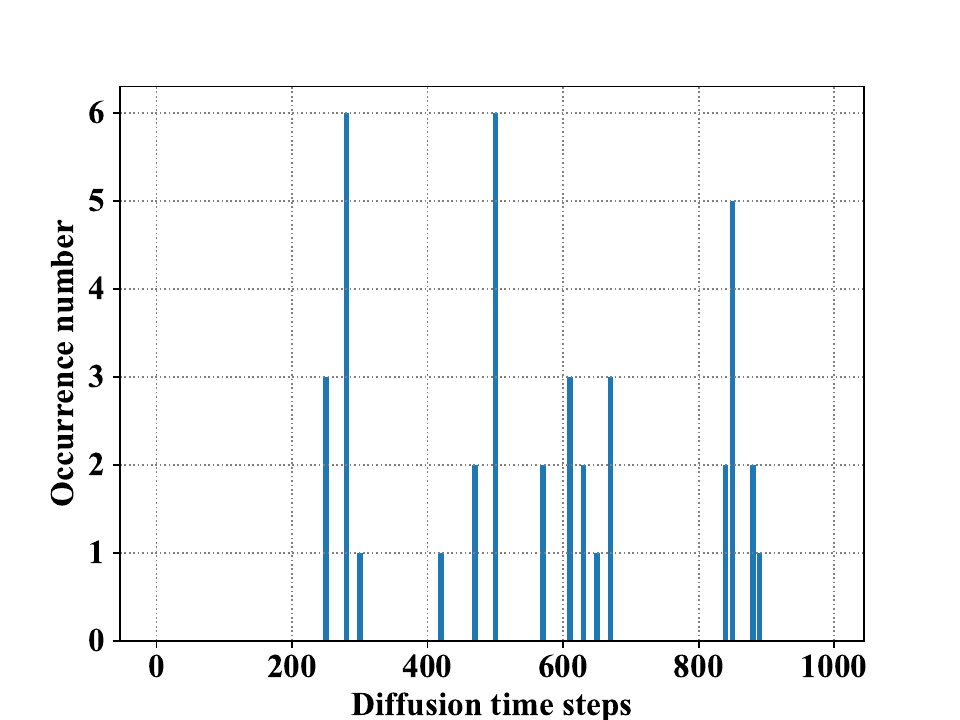}\label{Fig.migrate_adm_4_cfg7.5}}
    \caption{The occurrence number of time steps of top-10 candidates in Evolutionary search. 
    (a). Time steps occurrence number of ADM-G on ImageNet$64\times64$ with guidance scale 1.0. (b). Time steps occurrence number of ADM-G on ImageNet$64\times64$ with guidance scale 7.5. We observe that the distribution of occurrence number is changed depending on the guidance scale in generation process. }
\vspace{-1.2em}
\end{figure}
\begin{table*}
    \begin{center}
    \setlength{\tabcolsep}{3.5mm}{
    \begin{tabular}{ccccccc}
    \toprule[1pt]
    \makecell[c]{searched time steps} & \makecell[c]{searched model layers} & Steps & FID $\downarrow$ & IS $\uparrow$ & sampling time (s) & $N_\text{max}$\\ 
    \hline
      $\checkmark$ & $\checkmark$ & 4 & 14.53 & 38.24 & 4455 & 232 \\
      $\checkmark$ & $\times$ & 4 & 18.07 & 35.26 & 4476 & 232 \\
    \hline
      $\checkmark$ & $\checkmark$ & 6 & 10.26 & 45.35 & 6535 & 350 \\
      $\checkmark$ & $\times$ & 6 & 10.91 & 44.93 & 6712 & 350 \\
    \hline
      $\checkmark$ & $\checkmark$ & 10 & 6.08 & 54.62 & 10655 & 580 \\
      $\checkmark$ & $\times$ & 10 & 7.51 & 55.32 & 10719  & 580 \\
    \toprule[1pt]
    \end{tabular}
    }
    \end{center}
    \vspace{-0.8em}
    \caption{FID score and IS scores for ADM-G\cite{ADM} with our proposed method on ImageNet $64\times 64$ dataset. ``Sampling time (s)'' means the time to generate 50k samples. }
    \label{pruning}
    \vspace{-0.8em}
\end{table*}
\subsection{Migrate Search Results}
\label{sec.migrate}
We observe that the guidance scale in the generation process influences the search results significantly, and an optimal time steps sequence derived from one diffusion model can be transferred to another using the same guidance scale. Specifically, we search the optimal time steps sequence of length 4 for ADM-G on the ImageNet $64\times 64$ at guidance scales 1.0 and 7.5. The distribution of searched time steps for ADM-G with these guidance scales differ significantly, as shown in Fig.~\ref{Fig.migrate_adm_4} and Fig.~\ref{Fig.migrate_adm_4_cfg7.5}. 
Further, using a 7.5 guidance scale, we apply the optimal time steps of ADM-G on ImageNet $64\times 64$ to Stable Diffusion on COCO dataset, achieving an FID score of 24.11. In comparison, uniform time steps and the optimal time steps specifically searched for Stable Diffusion lead to FID scores of 38.25 and 20.93. This result suggests that we can obtain a desirable time steps sequence without repeating the search process when given a new diffusion model with the same guidance scale. However, we also find that applying the searched results from Stable Diffusion with guidance scale of 7.5 to ADM-G with guidance scale of 1.0 results in poor sample quality. This implies that the searched results from diffusion models with different guidance scales might not be transferable.

\subsection{Search for Time Steps and Architecture}
\label{sec.search_timesteps_and_arch}
We find that our method can achieve satisfactory performance when searching time steps only, but the performance can be further improved by searching model layers together with time steps. In this case, we constrain the sum of model layers at each time step to be less than $N_{\text{max}}$. 
We repeat the experiment under $N_{\text{max}} = 232$, $N_{\text{max}} = 350$, and $N_{\text{max}} = 580$, while the number of layers in noise prediction model is fixed to 58. After searching, we evaluate the FID score and IS score of diffusion models using the searched time steps and model layers. Besides, we also evaluate the performance of the diffusion model that only uses the searched time steps without using the searched model layers (\emph{e.g.} these diffusion models use a full noise prediction network to generate samples). In all these experiments, we don't retrain or fine-tune the searched subnet of the noise prediction network. 

Tab.~\ref{pruning} illustrates that the diffusion model with the searched model layers outperforms the model that employs a full noise prediction network in terms of both FID scores and generation speed. This result suggests that certain layers in the noise prediction network are superfluous.

We conduct an analysis on the searched architecture of Tab.~\ref{pruning}. We prune entire residual block and attention block from the noise prediction network in these experiments and observe that the importance of residual and attention blocks varies with the time-step length. Both residual and attention blocks are equally essential for the small time-step length, but attention blocks became increasingly important with more steps. 



\subsection{Comparison to the Prior Work}
We experiment with the DDPM provided by Alexander \emph{et al.} \cite{improved} on ImageNet $64\times 64$ against DDSS \cite{DDSS} which proposed to optimize the noise and time step schedule with differentiable diffusion sampler search. Tab.~\ref{comparison} demonstrates that our method can achieve a better FID score and IS score than DDSS. 

\begin{table}
    \begin{center}
    \begin{tabular}{cccc}
    \toprule[1pt]
    Method \textbackslash Steps & 5 & 10 & 15 \\
    \hline
    DDSS  & 55.14 / 12.9 & 37.32 /14.8 & 24.69 /17.2 \\
    \hline
    Ours  & 46.83 / 11.4 & 26.12 / 15.1 & 23.29 / 14.8 \\
    \hline
    \toprule[1pt]
    \end{tabular}
    \end{center}
    \vspace{-0.6em}
    \caption{FID score / IS score for our method against DDSS for the DDPM trained on ImageNet $64\times 64$ with $L_{\text{simple}}$ \cite{improved}}
\label{comparison}
\end{table}
\begin{table}[t]
    \begin{center}
    \setlength{\tabcolsep}{0.1mm}{
    \begin{tabular}{cccc}
    \toprule[2pt]
    Approach & Steps & \makecell[c]{Method Type} & \makecell[c]{Total Cost \\ (GPU days)} \\
    \hline
    \rule[0pt]{0pt}{10pt}
    \makecell[c]{AutoDiffusion } & 5 & Training-free Search & 1.125 \\
    \hline
    \rule[0pt]{0pt}{10pt}
    DDSS  & 5 & Reparameterization & 3.55 \\
    \hline
    \rule[0pt]{0pt}{10pt}
    Progressive Distil.(PD)  & 4 & Distillation & 359 \\
    \hline
    \rule[0pt]{0pt}{10pt}
    Progressive Distil.(PD)  & 8 & Distillation & 314 \\
    \toprule[2.2pt]
    \end{tabular}
    }
    \end{center}
    \vspace{-0.5em}
    \caption{Efficiency comparison. We assessed the computational resource demand of AutoDiffusion, PD, and DDSS using our reconstructed Improved-Diffusion codebase and ImageNet $64\times 64$ on a single V100 GPU. For DDSS, we approximated the computational resource consumption by running 50k training steps of U-Net and multiplying the training time by the time steps, as it executes the entire generation process in each training step.}
\label{GPU-days}
\end{table}

\subsection{The efficiency of AutoDiffusion}
AutoDiffusion is highly efficient and surpasses existing methods that demand additional computational resources such as PD \cite{progressive_kd} and DDSS \cite{DDSS} in computational resource requirements. AutoDiffusion uses a training-free search to determine time steps and diffusion models architecture, with search time depending on image resolution, time step length, and model size. Tab.~\ref{GPU-days} demonstrates the superior efficiency of AutoDiffusion compared to DDSS and PD. The computational resource required by DDSS and PD is approximately 3.15$\times$ and 279$\times$ that of AutoDiffusion.

\section{Conclusion}
In this paper, we propose AutoDiffusion to search the optimal time steps and architectures for any pre-trained diffusion models. We design a unified search space for both time steps and architectures, and then utilize the FID score as the evaluation metric for candidate models. We implement the evolutionary algorithm as the search strategy for the AutoDiffusion framework. Extensive experiments demonstrate that AutoDiffusion can search for the optimal time steps sequence and architecture with any given number of time steps efficiently. 
Designing more sophisticated methods that can evaluate the performance of diffusion models faster than FID score can improve the search speed and performance of AutoDiffusion, which we leave as future work. 

\textbf{Acknwoledgement. }
This work was supported by National Key R\&D Program of China (No.~2022ZD0118202), the National Science Fund for Distinguished Young Scholars (No.62025603), the National Natural Science Foundation of China (No.~U21B2037, No.~U22B2051, No.~62176222, No.~62176223, No.~62176226, No.~62072386, No.~62072387, No.~62072389, No.~62002305 and No.~62272401), and the Natural Science Foundation of Fujian Province of China (No.~2021J01002, No.~2022J06001).

{\small
\bibliographystyle{ieee_fullname}
\bibliography{egbib}
}

\begin{strip}
\null
\vskip .375in
\begin{center}
  {\Large \bf Appendix of ``AutoDiffusion: Training-Free Optimization of Time Steps and Architectures for Automated Diffusion Model Acceleration'' \par}
  \vspace*{24pt}
  {
  \large
  \lineskip .5em
  \par
  }
\end{center}
\end{strip}

\section*{A. Pseudo-code of Evolutionary Search}
The evolution algorithm utilized in our method is elaborated in Alg.~\ref{alg_ea}. Given a trained diffusion model, we sample candidates from search space randomly to form an initial population. During each iteration, we calculate FID score for each candidate in the population. After that, the Top $k$ candidates with the lowest FID score are selected as parents. We then apply cross and mutation to these parents to generate a new population for the next iteration. The aforementioned process is iteratively executed until the predetermined maximum number of iterations is attained.


\section*{B. Experiments details and more samples on Stable Diffusion}
\begin{figure*}
\begin{center}
\includegraphics[width=0.98\textwidth]{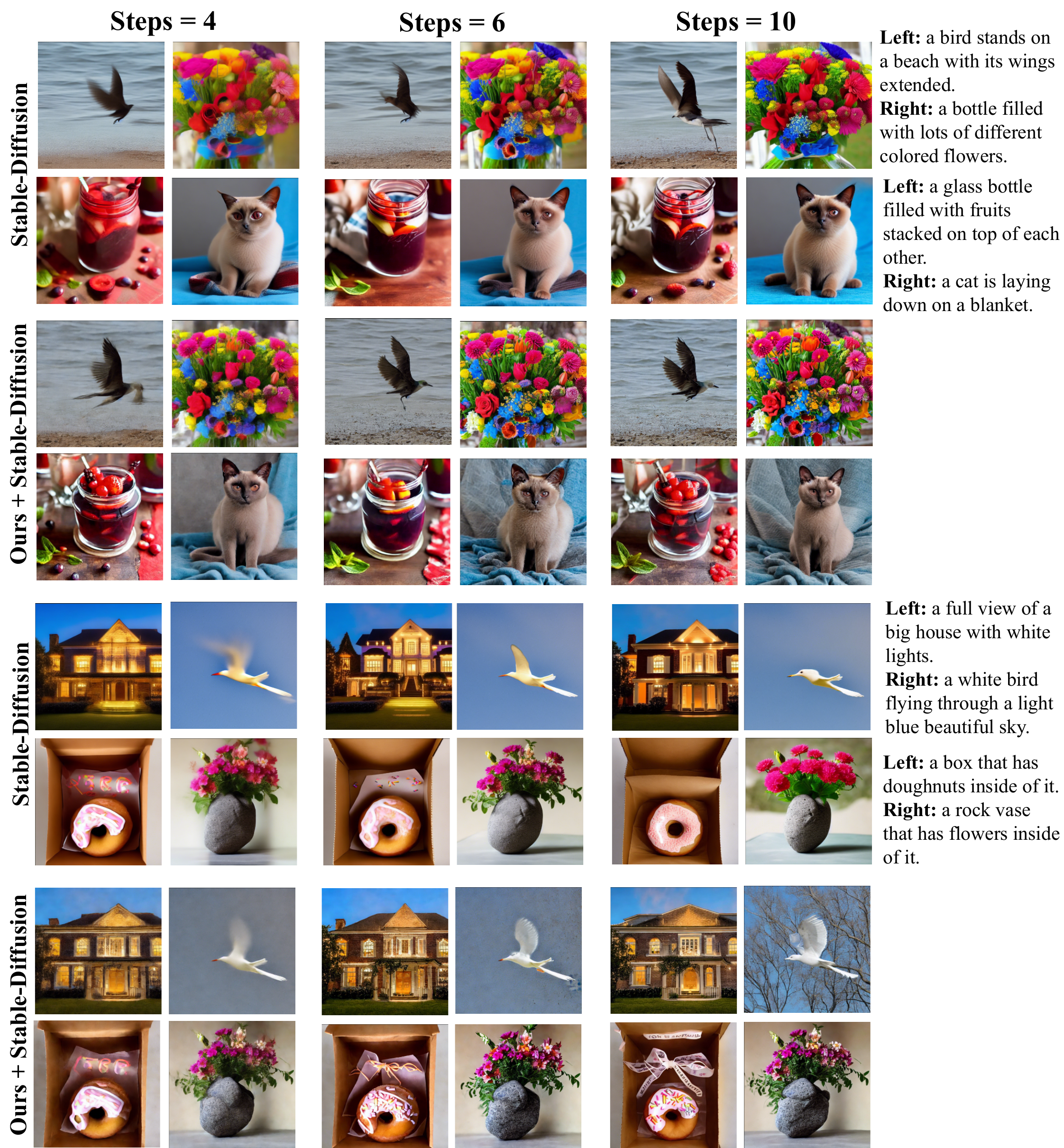}
\end{center}
   \caption{Samples obtained by Stable diffusion with and without our methods using the same random seed.}
\label{fig.sd_timesteps_supp}
\vspace{3cm}
\end{figure*}

\begin{figure*}
\begin{center}
\includegraphics[width=0.98\textwidth]{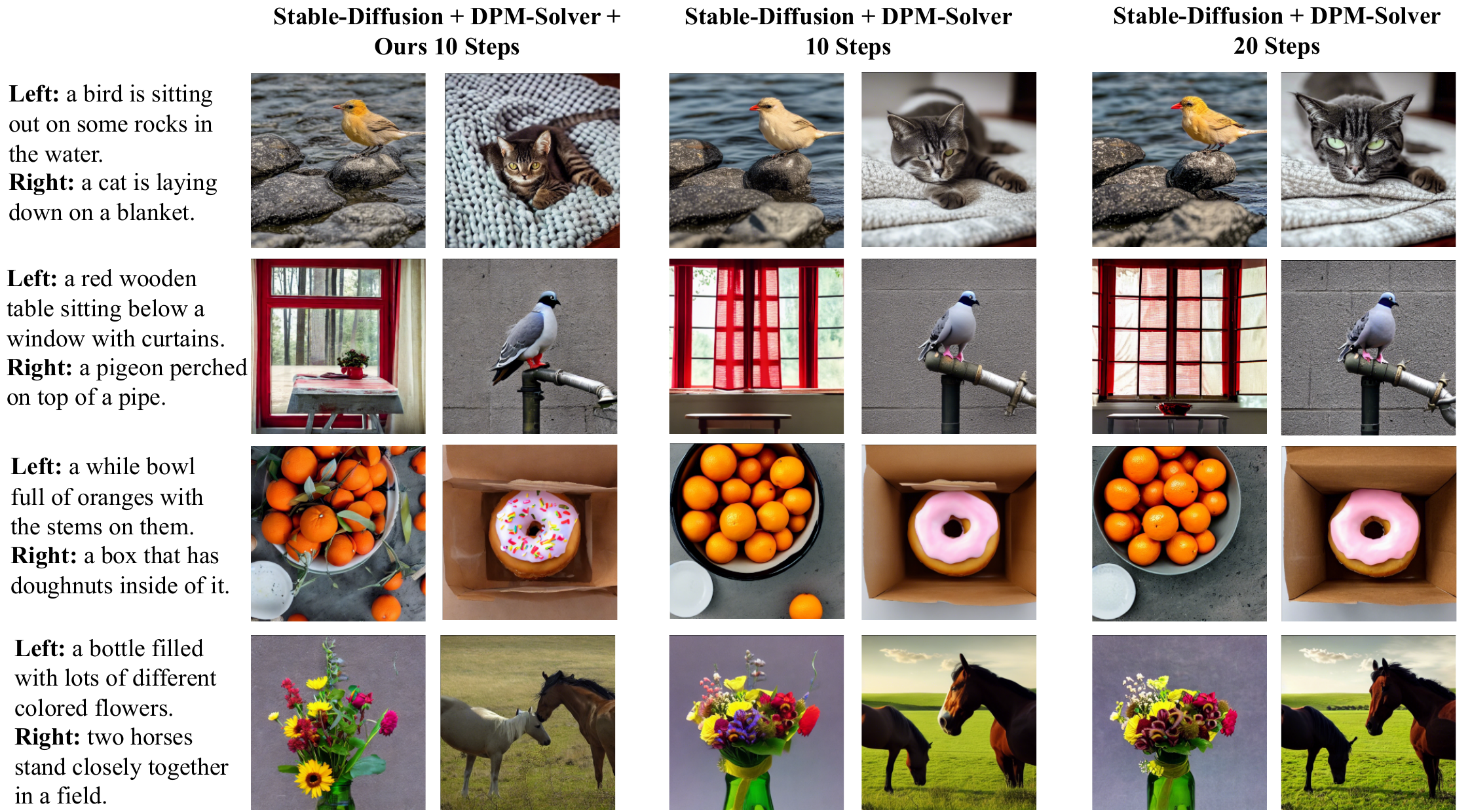}
\end{center}
   \caption{Samples generated by Stable Diffusion using DPM-Solver with our method at 10 steps are comparable to those generated only using DPM-Solver at 20 steps, and better than those generated only using DPM-Solver at 10 steps.}
\label{fig.sd_timesteps1}
\end{figure*}

For the experiments on Stable Diffusion \cite{LDM}, we utilize the official code and the released ``sd-v1-4.ckpt'' checkpoint\footnote{https://github.com/CompVis/stable-diffusion}. We employ the validation set of COCO 2014 dataset and 10k generated samples to obtain the FID score for Fig. 4 in the primary manuscript. And Tab.~\ref{SD} displays the detailed FID score corresponding to Fig. 4 in the primary manuscript.
Additional sampling results on Stable Diffusion using DPM-Solver \cite{dpmsolver} with and without our method are reported in Fig.~\ref{fig.sd_timesteps_supp} and Fig.~\ref{fig.sd_timesteps1}. 

\begin{figure*}
\begin{center}
\includegraphics[width=0.98\textwidth]{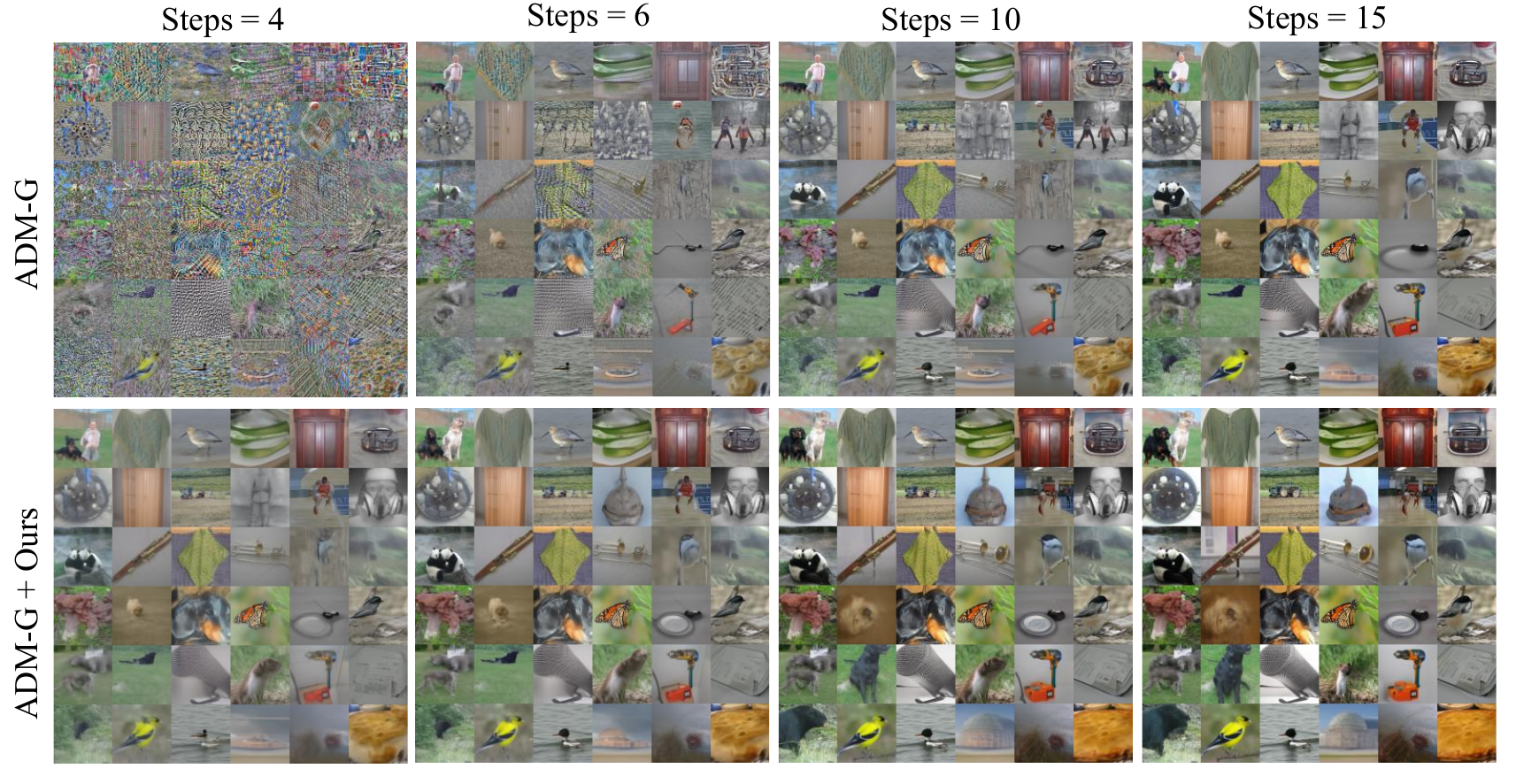}
\end{center}
   \caption{Samples generated by ADM pre-trained on ImageNet $64\times 64$ cat with and without our method.}
\label{fig.ImageNet}
\vspace{1cm}
\end{figure*}

\begin{table*}
    \begin{center}
    \begin{tabular}{cccccccc}
    \toprule[1pt]
    Ours & Steps & \makecell[c]{FID $\downarrow$ \\ DPM-Solver } & \makecell[c]{IS $\uparrow$ \\ DPM-Solver } & \makecell[c]{FID $\downarrow$ \\ DDIM } & \makecell[c]{IS $\uparrow$ \\ DDIM } & \makecell[c]{FID $\downarrow$ \\ PLMS } & \makecell[c]{IS $\uparrow$ \\ PLMS } \\
    
    \toprule[1pt]
    $\times$  & 4 & 22.43  & 29.70 & 39.13  & 23.05 & 38.22  & 22.00 \\
    $\checkmark$ & 4 & 18.22 (-4.21) & 33.10 (+3.40) & 26.72 (-12.41) & 27.73 (+4.68) & 20.94 (-17.28) & 30.38 (+8.38)  \\
    \hline
    $\times$  & 6 & 17.36 & 34.03 & 18.87 & 31.63 & 32.40 & 24.41 \\
    $\checkmark$ & 6 & 12.95 (-4.41) & 34.26 (+0.23) & 16.44 (-2.43) & 33.70 (+2.07) & 16.48 (-15.92) & 33.58 (+9.17)  \\
    \hline
    $\times$  & 10 & 15.95 & 36.23  & 14.93 & 34.97 & 19.16 & 30.22 \\
    $\checkmark$ & 10 & 12.67 (-3.28) & 36.54 (+0.31) & 14.06 (-0.87) & 35.38 (+0.42) & 13.57 (-5.59) & 36.79 (+6.57)  \\
    \toprule[1pt]
    \end{tabular}
    \end{center}
    \caption{FID score and IS scores for Stable Diffusion using DPM-Solver \cite{dpmsolver}, DDIM \cite{ddim} and PLMS \cite{plms} with and without our method on COCO dataset, varying the number of time steps.}
    \label{SD}
\end{table*}

\section*{C. Experiments details and more samples on ADM}
\begin{figure*}
\begin{center}
\includegraphics[width=0.95\textwidth]{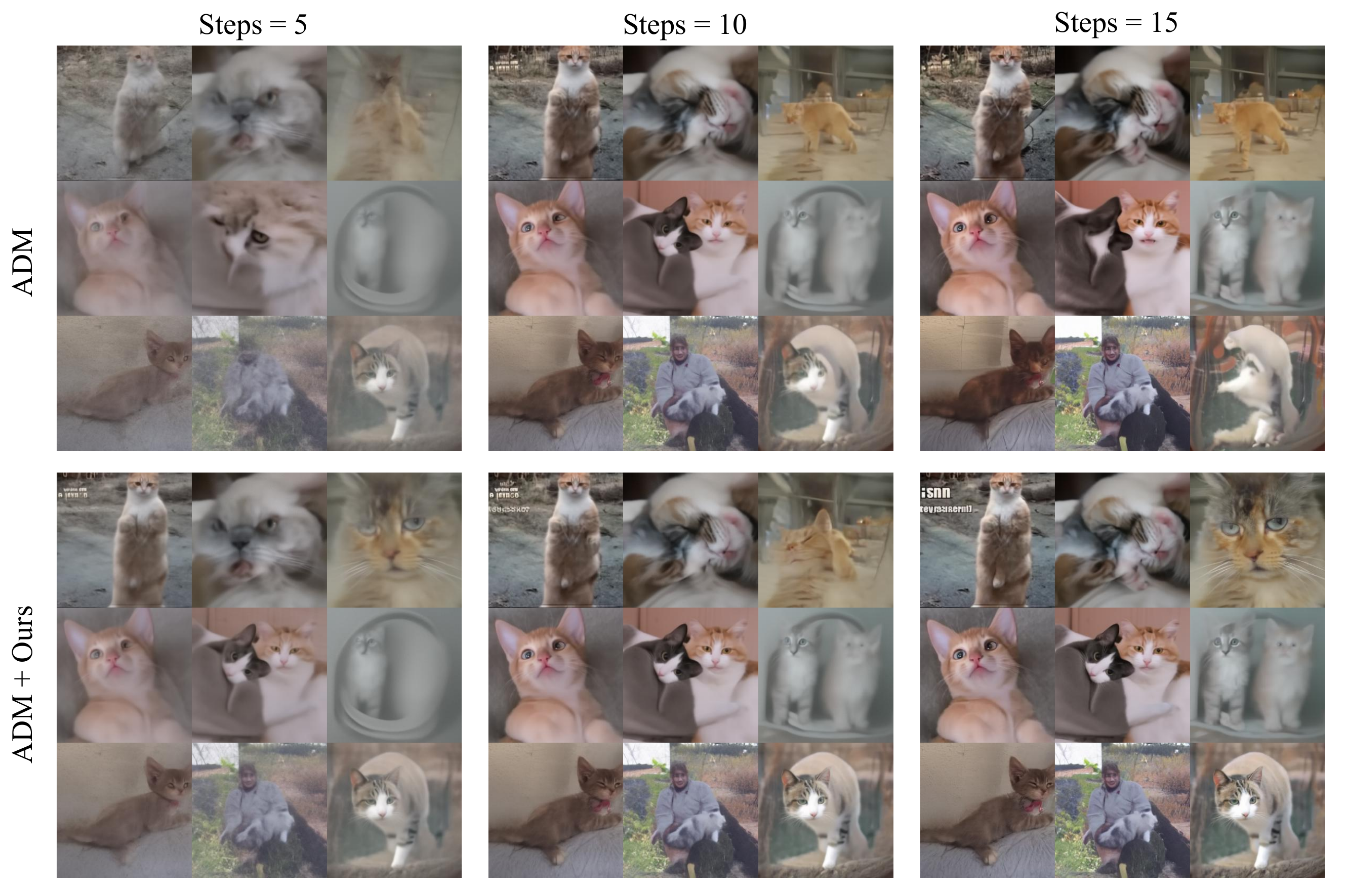}
\end{center}
   \caption{Samples generated by ADM pre-trained on LSUN cat with and without our method.}
\label{fig.LSUNcat}
\end{figure*}

We use the official code and released checkpoint\footnote{https://github.com/openai/guided-diffusion} for the experiments with ADM-G and ADM \cite{ADM} on ImageNet, LSUN cat, and LSUN bedroom. In these experiments, we utilize 50k generated images and pre-computed sample batches from the reference datasets available in the codebase \footnote{https://github.com/openai/guided-diffusion/tree/main/evaluations} of ADM to calculate the FID score of Tabs 2 and 3 in our main manuscript. Additional sampling results on ImageNet $64\times 64$ and LSUN cat are reported in Fig.~\ref{fig.ImageNet} and Fig.~\ref{fig.LSUNcat}, respectively.

\begin{algorithm}[t]
    \caption{Evolutionary search}
    \renewcommand{\algorithmicrequire}{\textbf{Input:}}
    \renewcommand{\algorithmicensure}{\textbf{Output:}}
    \label{alg_ea}
    \begin{algorithmic}[1]
    \REQUIRE Pre-trained diffusion model $D$, Number of searched time steps $K$, population size $P$, max iteration $MaxIter$, mutation probability $p$, the number of candidate generated from cross $N_{c}$ and mutation $N_m$.
    \ENSURE The best candidate $cand^*$  
    
    \STATE {$\text{P}_0 = InitializePopulation(P)$}
    \STATE {$\text{Topk} = \emptyset $}
    \FOR{ $i = 1: MaxIter$}
        \STATE{Samples = $GenerationProcess(D, \text{P}_{i-1})$}
        \STATE{ $\text{FID}_{i-1} = CalculateFID(\text{Samples})$}
        \STATE{ Topk = $UpdataTopk(\text{Topk}, \text{P}_{i-1}, \text{FID}_{i-1})$}
        \STATE{ $\text{P}_{\text{cross}} = cross(\text{Topk}, N_c)$}
        \STATE{ $\text{P}_{\text{mutation}} = mutation(\text{Topk}, N_m, p)$}
        \STATE{ $\text{P}_i = \text{P}_{\text{cross}} + \text{P}_{\text{mutation}}$}
    \ENDFOR
    \STATE{ $cand^* = Top1(\text{Topk})$}
    \end{algorithmic}
\end{algorithm}

\begin{table*}
    \begin{center}
    \setlength{\tabcolsep}{4mm}{
    \begin{tabular}{cccc}
    \toprule[1pt]
    Performance Estimation Strategy \textbackslash Steps & 4 & 6 & 10 \\
    \hline
    FID score  & 17.86 / 34.88 & 11.17 / 43.47 & 6.24 / 57.85 \\
    \hline
    KID score  & 21.06 / 30.78 & 12.68 / 39.42 & 9.72 / 42.60 \\
    \hline
    KL-divergence  & 414.9 / 1.125 & 414.3 / 1.13 & 414.8 / 1.14 \\
    \toprule[1pt]
    \end{tabular}
    }
    \end{center}
    \caption{FID score / IS score for the performance estimation ablation on ImageNet $64\times 64$.}
\label{ablation_on_FID}
\end{table*}

\begin{table*}
\setlength{\tabcolsep}{8mm}{
    \begin{center}
    \begin{tabular}{cccc}
    \toprule[1pt]
    Method \textbackslash Steps & 4 & 6 & 10 \\
    \hline
    Evolutionary Search  & 17.86 / 34.88 & 11.17 / 43.47 & 6.24 / 57.85 \\
    \hline
    Random Search  & 18.84 / 34.17 & 11.17 / 43.02 & 7.05 / 51.43 \\
    \hline
    Uniform Time steps  & 138.66 / 7.06 & 23.71 / 31.53 & 8.86 / 46.50 \\
    \toprule[1pt]
    \end{tabular}
    \end{center}
    }
    \caption{FID score / IS score for the search algorithm ablation on ImageNet $64\times 64$.}
\label{ablation_on_ea}
\end{table*}

\begin{table*}[h!]
\setlength{\tabcolsep}{4mm}{
    \begin{center}
    \begin{tabular}{ccc}
    \toprule[1pt]
    \textbf{Diffusion Models} & \textbf{Dataset} & \textbf{Optimal Time Steps} \\
    \hline
    ADM-G + DDIM & ImageNet $64\times 64$ & [926, 690, 424, 153]\\
    \hline
    Stable Diffusion + PLMS & COCO & [848, 598, 251, 21] \\
    \hline
    Stable Diffusion + DPM-Solver & COCO & [0.9261, 0.7183, 0.5005, 0.2857, 0.0150] \\
    \toprule[1pt]
    \end{tabular}
    \end{center}
    }
    \caption{Optimal time steps sequence with length 4 for different diffusion models.}
\label{optimal_timesteps4}
\end{table*}

\begin{table*}[h!]
\setlength{\tabcolsep}{2mm}{
    \begin{center}
    \begin{tabular}{ccc}
    \toprule[1pt]
    \textbf{Diffusion Models} & \textbf{Dataset} & \textbf{Optimal Time Steps} \\
    \hline
    ADM-G + DDIM & ImageNet $64\times 64$ & [123, 207, 390, 622, 830, 948]\\
    \hline
    Stable Diffusion + PLMS & COCO & [19, 130, 335, 519, 695, 931] \\
    \hline
    Stable Diffusion + DPM-Solver & COCO &  [0.9261, 0.6670, 0.5005, 0.3340,  0.1548, 0.0150, 0.0120] \\
    \toprule[1pt]
    \end{tabular}
    \end{center}
    }
    \caption{Optimal time steps sequence with length 6 for different diffusion models}
\label{optimal_timesteps6}
\end{table*}

\begin{table*}[h!]
    \begin{center}
    \setlength{\tabcolsep}{7mm}{
    \begin{tabular}{ccc}
    \toprule[1pt]
    Index of removed model layers & Steps & $N_\text{max}$\\ 
    \hline
      \{[], [], [], [55]\} & 4 & 232 \\

  \hline
  \{[], [], [], [], [], [52]\} & 6 & 350 \\

  \hline
  \{[], [], [], [], [], [], [30, 10, 39, 4, 15, 46, 49, 54, 8], [], [], []\} & 10 & 580 \\
      
    \toprule[1pt]
    \end{tabular}
    }
    \end{center}
    \caption{Index of removed model layers in the optimal architecture searched for ADM-G on ImageNet $64\times 64$. ``[]'' means no layer is removed at corresponding time step.}
    \label{model_layers}
\end{table*}

\section*{D. Ablation Study}
\subsection*{D.1. Ablation on Performance Estimation}
To assess the impact of performance estimation, we conduct experiments employing different evaluation metrics. Specifically, we replicate the experiment on ImageNet $64 \times 64$ with ADM-G using FID score, KID score, and KL-divergence as performance estimation. In these experiments, we only focus on time step optimization and use a complete noise prediction network. The results summarized in Tab.~\ref{ablation_on_FID} indicate that there is little difference in the performance of FID score and KID score. This observation can be attributed to the fact that both FID score and KID score gauge the distance between the statistical properties of the feature of generated samples and real samples. In contrast, the performance of KL-divergence is poor, which demonstrates that KL-divergence is inadequate in estimating the performance of the time steps sequence properly. 

\subsection*{D.2. Ablation on Search Algorithm}
We conduct experiments to examine the impact of various search algorithms on experimental results. Specifically, we utilize evolutionary search and random search to search the optimal time steps sequence for ADM-G on ImageNet $64\times 64$. The results presented in Tab.~\ref{ablation_on_ea} illustrate that the selection of the search algorithm does not significantly influence the experimental results. Notably, We observe that even the time steps sequence searched by the simplistic random search algorithm produces better sample quality than the uniform time steps sequence. 

\section*{E. Search Results}
\subsection*{E.1. Time Steps Sequence}
\begin{figure*}
    \centering
    \subfigure[]{\includegraphics[width=0.32\textwidth]{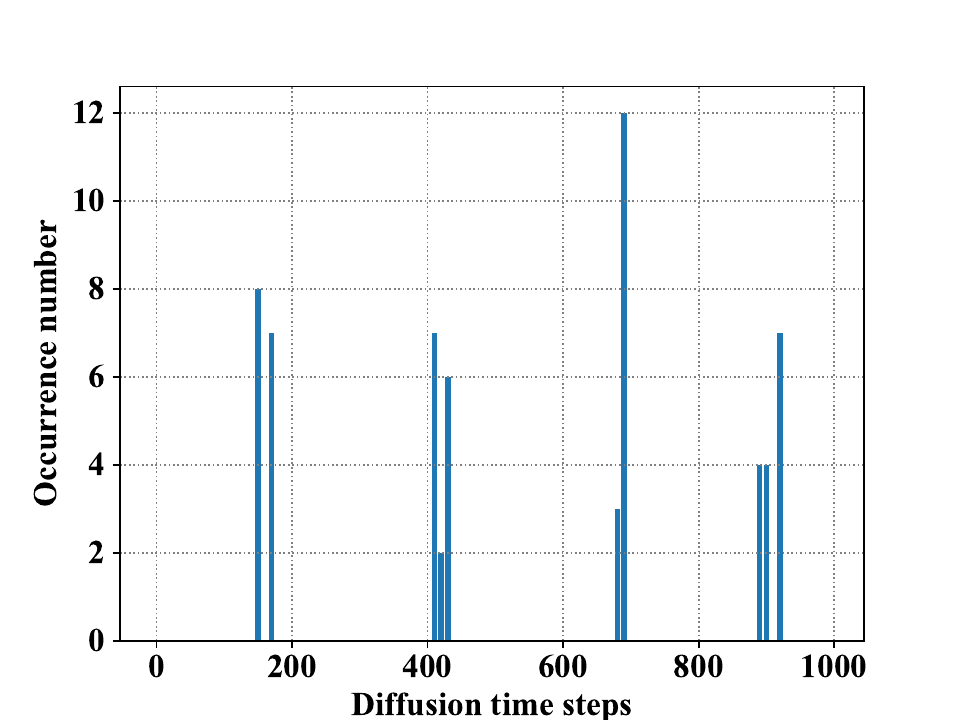}\label{Fig.I4}}
    \subfigure[]{\includegraphics[width=0.32\textwidth]{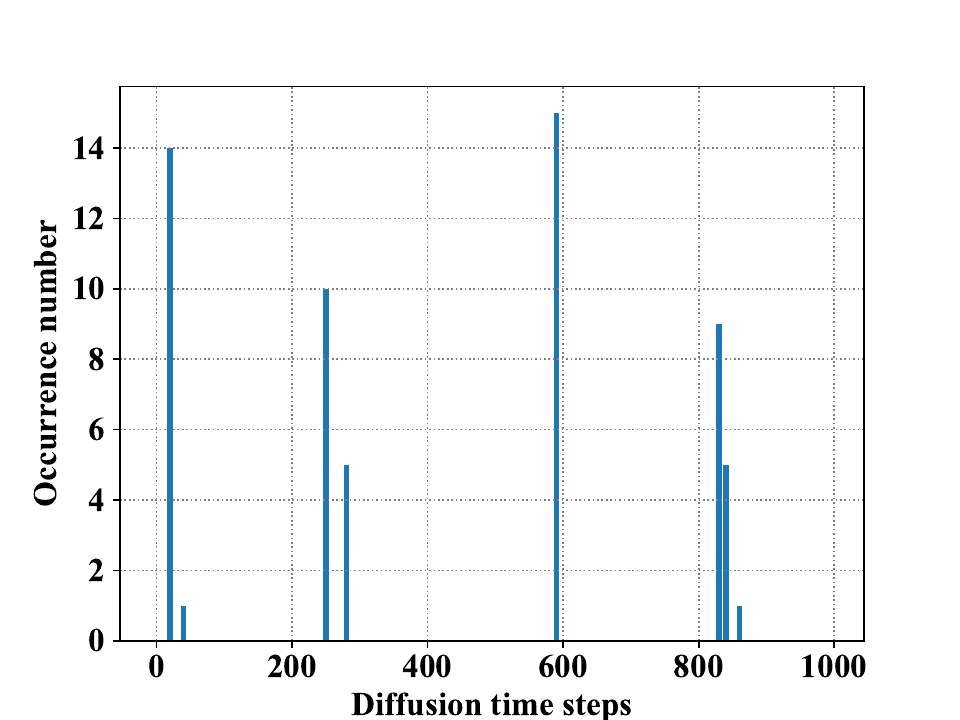}\label{Fig.P4}}
    \subfigure[]{\includegraphics[width=0.32\textwidth]{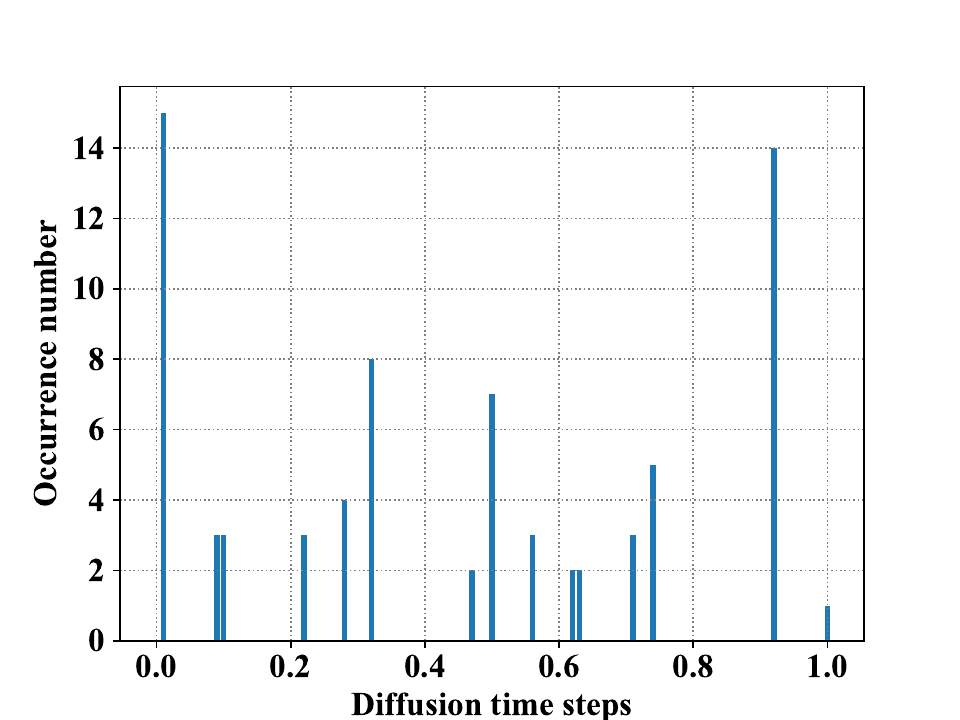}\label{Fig.D4}}
    \subfigure[]{\includegraphics[width=0.32\textwidth]{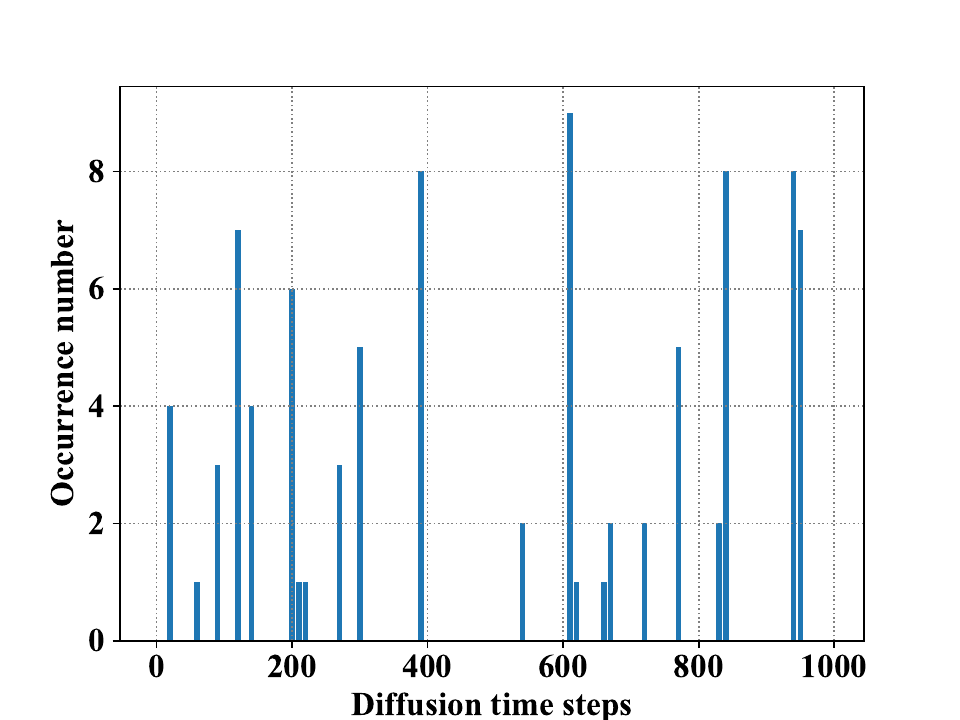}\label{Fig.I6}}
    \subfigure[]{\includegraphics[width=0.32\textwidth]{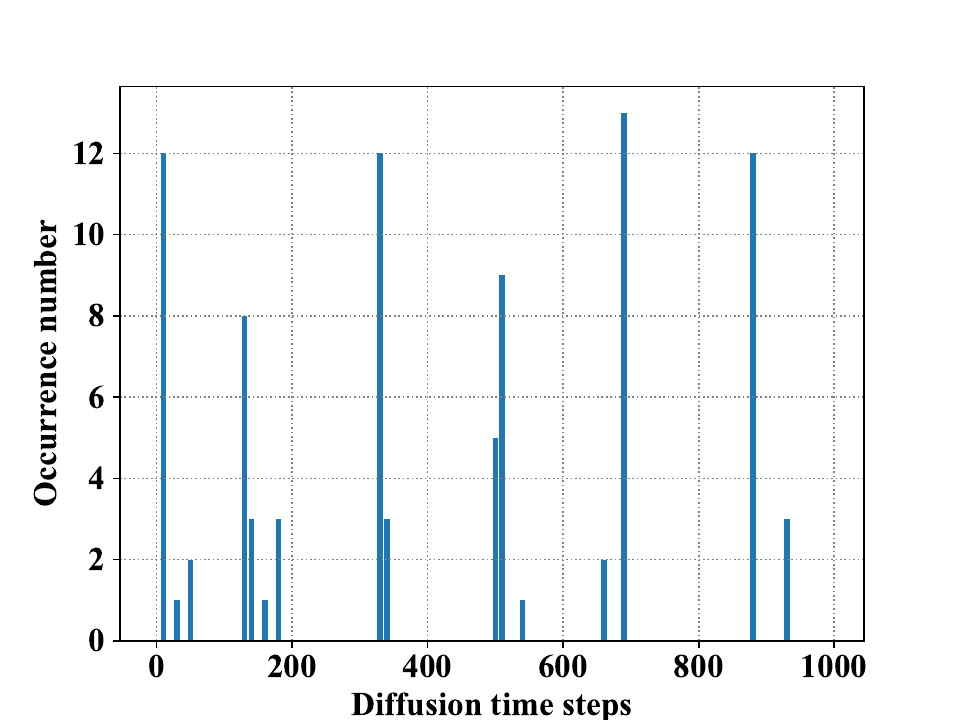}\label{Fig.P6}}
    \subfigure[]{\includegraphics[width=0.32\textwidth]{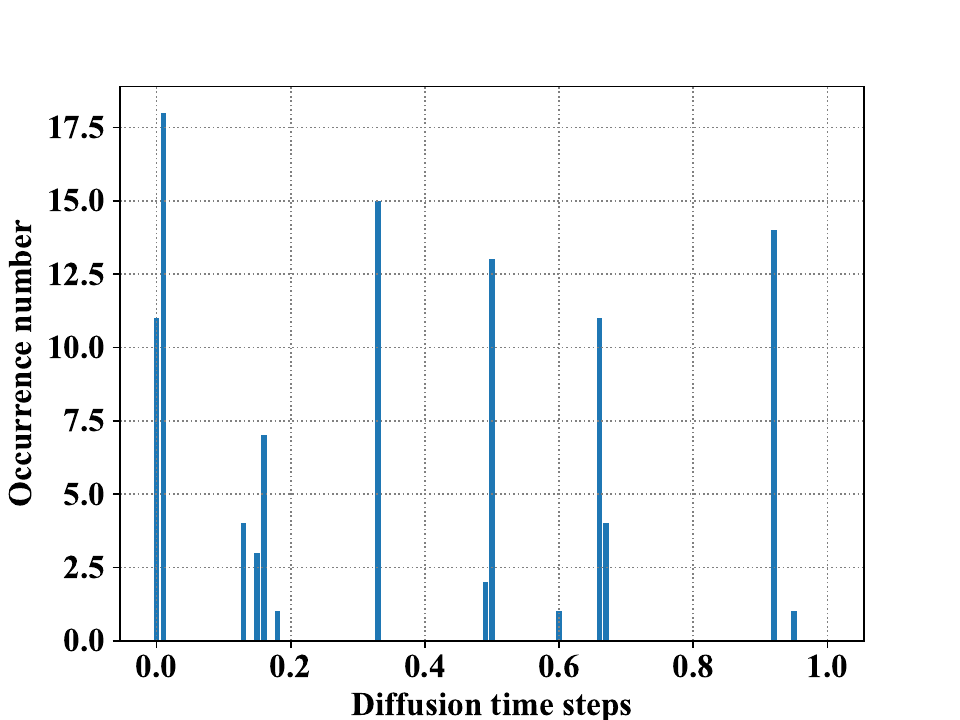}\label{Fig.D6}}
    \caption{The occurrence number of time steps of top-15 candidates in Evolutionary search. (a). Occurrence number of time steps in the top-15 sequence with length 4 for ADM-G using DDIM on ImageNet$64\times64$. (b). Occurrence number of time steps in the top-15 sequence with length 4 for Stable Diffusion using PLMS on COCO dataset. (c). Occurrence number of time steps in the top-15 sequence with length 4 for Stable Diffusion using DPM-Solver on COCO dataset. (d). Occurrence number of time steps in the top-15 sequence with length 6 for ADM-G using DDIM on ImageNet$64\times64$. (e). Occurrence number of time steps in the top-15 sequence with length 6 for Stable Diffusion using PLMS on COCO dataset. (f). Occurrence number of time steps in the top-15 sequence with length 6 for Stable Diffusion using DPM-Solver on COCO dataset.}
    \label{Fig.searched_timesteps}
\end{figure*}

The optimal time step sequences in the evolutionary search for different diffusion models are presented in Tab.~\ref{optimal_timesteps4} and Tab.~\ref{optimal_timesteps6}. Besides, Fig.~\ref{Fig.searched_timesteps} illustrates the occurrence number of time steps of the top-15 candidates in the evolutionary search. In these experiments, the max time step of Stable Diffusion with DPM-Solver is 1, while the other diffusion models are 1000. When searching the optimal time steps for Stable Diffusion with DPM-Solver, we follow the strategy of DPM-Solver that uses the time steps sequence with a length of (Steps + 1)\footnote{https://github.com/CompVis/stable-diffusion/blob/main/ldm/models\\/diffusion/dpm\_solver/dpm\_solver.py}. We observe that the optimal time steps tend to cluster within a specific interval. In addition, the distribution of these optimal time steps markedly differs between ADM-G and Stable Diffusion due to their distinct guidance scales.

\subsection*{E.2. Model Architectures}
The optimal architecture layers in the evolutionary search for ADM-G on ImageNet $64\times 64$ are shown in Tab.~\ref{model_layers}. In these experiments, we number each layer of the complete noise prediction network ascending from the input layer to the output layer. As described in the main manuscript, we constrain the sum of model layers at each time step to be less than $N_\text{max}$. And the complete noise prediction network comprises 58 model layers. We observe that the number of removed model layers is higher when $N_\text{max} = 580$ compared to $N_\text{max} = 232$ and $N_\text{max} = 350$. This observation highlights an increase in model redundancy with an increase in the number of time steps.

\end{document}